%% file: main.tex
\documentclass[runningheads]{llncs}

\usepackage{eccv}

\usepackage{eccvabbrv}

\usepackage{graphicx}
\usepackage{booktabs}
\usepackage{glossaries}
\usepackage{wrapfig}
\usepackage{multicol, multirow}

\usepackage[accsupp]{axessibility}  

\usepackage{hyperref}

\usepackage{orcidlink}

\input{preamble}

\newacronym{ai}{AI}{Artificial Intelligence}
\newacronym{cv}{CV}{Computer Vision}
\newacronym{cnn}{CNN}{Convolutional Neural Network}
\newacronym{ml}{ML}{Machine Learning}
\newacronym{mlr}{MLR}{Multinomial Logistic Regression}
\newacronym{dl}{DL}{Deep Learning}
\newacronym{ood}{OOD}{Out-Of-Distribution}
\newacronym{id}{ID}{In-Distribution}
\newacronym{hypo}{HYPO}{Hyperbolic Outlier Synthesis}
\newacronym{npos}{NPOS}{Non-parametric Outlier Synthesis}
\newacronym{vos}{VOS}{Virtual Outlier Synthesis}
\newacronym{erm}{ERM}{Empirical Risk Minimization}
\newacronym{fpr95}{FPR95}{False Positive Rate at 95\% recall}
\newacronym{auroc}{AUROC}{Area Under the Receiver Operating Characteristic Curve}
\newacronym{knn}{KNN}{K-Nearest Neighbors}

\begin{document}


\title{Hyperbolic Metric Learning for Visual Outlier Detection}


\author{Alvaro Gonzalez-Jimenez\inst{3}\orcidlink{0000-0002-1337-9430} \and
	Simone Lionetti\inst{2}\orcidlink{0000-0001-7305-8957} \and
	Dena Bazazian\inst{1}\orcidlink{0000-0002-1229-4494} \and
	Philippe Gottfrois\inst{3}\orcidlink{0000-0001-8023-3207} \and
	Fabian Gröger\inst{2,3}\orcidlink{0000-0002-9699-688X} \and
	Alexander Navarini\inst{3}\orcidlink{0000-0001-7059-632X} \and
	Marc Pouly\inst{2}\orcidlink{0000-0002-9520-4799}
}

\authorrunning{Gonzalez-Jimenez \etal}

\institute{
	University of Plymouth, Plymouth, England \\
	\email{dena.bazazian@plymouth.ac.uk} \\ \and
	Lucerne School of Computer Science and Information Technology, Rotkreuz, Switzerland\\
	\email{\{name.lastname\}@hslu.ch}\\ \and
	University of Basel, Basel, Switzerland\\
	\email{\{name.lastname\}@unibas.ch}
}

\maketitle

\begin{abstract}
	Out-Of-Distribution (OOD) detection is critical to deploy deep learning models in safety-critical applications.
	OOD typically works by minimising intra-class and maximising inter-class separation, which can profit from the exponential volume growth in Hyperbolic space. This work proposes a metric framework that leverages the strengths of Hyperbolic geometry for OOD detection. Inspired by previous works that refine the decision boundary for OOD data with synthetic outliers, we extend this method to Hyperbolic space. Interestingly, we find that synthetic outliers do not benefit OOD detection in Hyperbolic space as they do in Euclidean space. Furthermore we explore the relationship between OOD detection performance and Hyperbolic embedding dimension, addressing practical concerns in resource-constrained environments. Extensive experiments show that our framework improves the FPR95 for OOD detection from 21\% to 16\% and from 49\% to 28\% on {CIFAR-10} and {CIFAR-100} respectively compared to Euclidean methods.

\end{abstract}

\input{content/introduction}
\input{content/preliminaries}

\input{content/methodology}
\input{content/experiments}
\input{content/literature}
\input{content/limitations}
\input{content/conclusions}

%
%
\bibliographystyle{splncs04}
\bibliography{egbib}

\appendix
\input{content/appendix}
\end{document}

%% file: preamble.tex
\usepackage[dvipsnames]{xcolor}

\renewcommand{\phi}{\varphi}
\renewcommand{\emptyset}{\varnothing}

\renewcommand{\tilde}[1]{\widetilde{#1}}

\newcommand{\ob}{\mathbf{o}}
\newcommand{\pb}{\mathbf{p}}

\newcommand{\sbb}{\mathbf{s}}

\newcommand{\ub}{\mathbf{u}}
\newcommand{\vb}{\mathbf{v}}
\newcommand{\wb}{\mathbf{w}}
\newcommand{\xb}{\mathbf{x}}
\newcommand{\yb}{\mathbf{y}}
\newcommand{\zb}{\mathbf{z}}

\newcommand{\Ib}{\mathbf{I}}

\newcommand{\Ob}{\mathbf{O}}

\newcommand{\Ccal}{\mathcal{C}}

\newcommand{\Gcal}{\mathcal{G}}

\newcommand{\Lcal}{\mathcal{L}}

\newcommand{\Pcal}{\mathcal{P}}

\newcommand{\Scal}{{\mathcal{S}}}
\newcommand{\Tcal}{{\mathcal{T}}}

\newcommand{\Vcal}{\mathcal{V}}
\newcommand{\Wcal}{\mathcal{W}}
\newcommand{\Xcal}{\mathcal{X}}
\newcommand{\Ycal}{\mathcal{Y}}

\newcommand{\E}{\mathbb{E}} 
\newcommand{\Lorentz}{\mathbb{L}} 
\newcommand{\Probability}{\mathbb{P}} 
\newcommand{\R}{\mathbb{R}} 
\def\iid{{\it i.i.d.}}

\newcommand{\inner}[2]{\left\langle #1,#2 \right\rangle}
\newcommand{\rbr}[1]{\left(#1\right)}
\newcommand{\sbr}[1]{\left[#1\right]}
\newcommand{\cbr}[1]{\left\{#1\right\}}

%% file: content/introduction.tex
\section{Introduction}
\label{sec:introduction}
The field of \gls{dl} has witnessed remarkable advancements, revolutionizing various domains from image recognition to natural language processing. While \gls{dl} models excel within their training data distribution, a critical challenge arises when encountering unforeseen data. This can lead to overly confident predictions with potentially disastrous real-world consequences \cite{hendrycksBaselineDetectingMisclassified2018}.

\gls{ood} detection has emerged as a pivotal research area aimed at addressing this challenge. Its objective is the identification of inputs that deviate significantly from the training data distribution. This enables the safe deployment of \gls{dl} models in scenarios where encountering novel or unexpected data is inevitable, such as medical imaging and autonomous vehicles, where erroneous predictions can have life-threatening implications.

Existing \gls{ood} detection approaches suffer from overconfidence on \gls{ood}, \ie data with higher likelihood leading to unreliable uncertainty estimates \cite{nguyenDeepNeuralNetworks2015a}, rely on the quality of feature embeddings \cite{mingHowExploitHyperspherical2023, sunOutofdistributionDetectionDeep2022} or
require the use of external auxiliary
\gls{ood} data \cite{bevandicDiscriminativeOutofdistributionDetection2018, malininPredictiveUncertaintyEstimation2018, geifmanSelectiveNetDeepNeural2019, heinWhyReluNetworks2019a, meinkeNeuralNetworksThat2020a, jeongOODMAMLMetalearningFewshot2020, yangSemanticallyCoherentOutofdistribution2021a}, which poses challenges especially when the exact nature of the \gls{ood} data is unknown. Synthesizing outliers from the low-likelihood feature space of \gls{id} data has emerged as an alternative to obtain this auxiliary data \cite{duVOSLearningWhat2022, taoNonparametricOutlierSynthesis2022}.
However, this assumes that \gls{ood} samples lie on the periphery of the \gls{id} distribution, which may not always be true. As a result, synthesized outliers may not accurately reflect the characteristics of true \gls{ood} data, leading to suboptimal detection performance.

One common ingredient that all the previously mentioned frameworks share is that they are based on \textit{Euclidean} geometry. In this paper, we argue that real-world visual data often exhibit intricate hierarchical structures, such as the organization of diseases in medical imaging.  However, Euclidean geometry fails to accurately capture these hierarchical relationships, leading to suboptimal representations \cite{bridsonMetricSpacesNonpositive1999}. Herein lies our motivation for exploring \textit{Hyperbolic} geometry as an alternative.

In Hyperbolic geometry, unlike in Euclidean geometry, given a line and a distinct point, there are two lines parallel to the first one that pass through the point. This distinction defines Hyperbolic space as a framework of constant negative curvature, known for its hierarchical organization and exponential expansion properties \cite{bridsonMetricSpacesNonpositive1999}.
Recent interest in Hyperbolic space for \gls{cv} has spurred significant advancements in embedding complex hierarchical structures with minimal distortion \cite{salaRepresentationTradeoffsHyperbolic2018, ganeaHyperbolicNeuralNetworks2018}.
This offers distinct advantages across various domains, including few-shot learning \cite{pengHyperbolicDeepNeural2021, khrulkovHyperbolicImageEmbeddings2020}, unsupervised learning \cite{parkUnsupervisedHyperbolicRepresentation2021}, and representation learning \cite{desaiHyperbolicImagetextRepresentations2024, ganeaHyperbolicEntailmentCones2018, lawLorentzianDistanceLearning2019}.
Moreover, previous research hints at the potential of Hyperbolic space for \gls{ood} detection, leveraging uncertainty measurements based on the distance of the embeddings from the origin in the Poincar\'e disk \cite{khrulkovHyperbolicImageEmbeddings2020, ghadimiatighHyperbolicImageSegmentation2022} or using the energy-based score \cite{vanspenglerPoincarResNet2023, guoClippedHyperbolicClassifiers2022}. However, model performance is yet not competitive with Euclidean-based models.

We hypothesize that the gap between Euclidean and Hyperbolic space arises from differences in the training objectives employed, particularly concerning \textit{intra-class} variation and \textit{inter-class} separation. Intra-class variation evaluates the consistency of representations across diverse samples of the same class, whereas inter-class separation gauges the spread of characteristics across samples of distinct classes. Ideally, features should exhibit minimal variation within classes and significant separation between classes to effectively detect \gls{ood} data.

In this paper, we propose HOD (\textbf{H}yperbolic \textbf{O}utlier \textbf{D}etection), a strategy that leverages the Hyperbolic space to learn \gls{id} representations promoting low variations and high separation. Our learning objective establishes a new state-of-the-art in \gls{ood} detection, with an improvement in the \gls{fpr95} from 21\% to 16\%, and from 49\% to 28\% on {CIFAR-10} and {CIFAR-100} respectively compared to current Euclidean methods. Through empirical evaluations, we demonstrate the effectiveness of certain techniques to detect \gls{ood} in the Hyperbolic space while identifying others that are not suitable. Furthermore, we propose the first strategy for synthesizing outliers in the Hyperbolic space, and show that it does not significantly improve results. We demonstrate that Hyperbolic space is better suited for deployment in real-world scenarios, particularly in resource-constrained environments which require low-dimensional embeddings. This opens the door for many practical applications in various industrial domains.

\begin{figure}[t!]
	\centering
	\includegraphics[width=\linewidth]{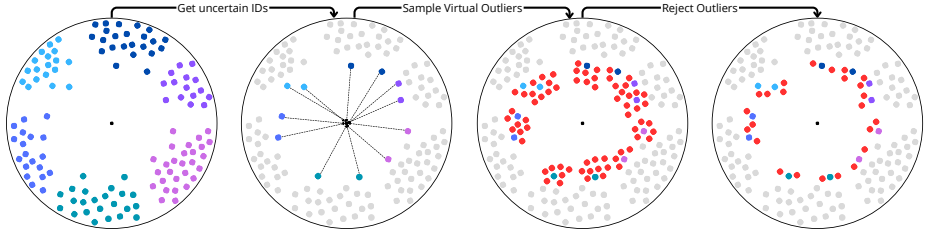}
	\caption{Illustration of the Hyperbolic outliers synthesis. First, the \gls{id} embeddings are optimized using \cref{eq:Hyperbolic_contrastive_loss} which encourages low intra-class variation and high inter-class separation. Second, the most uncertain \gls{id} embeddings are selected by their distance towards the origin. Third, the outliers are synthesized by sampling from the wrapped Gaussian distribution centered around the selected \gls{id} embeddings. Finally, only those with a higher level of certainty are kept.}
	\label{fig:sampling-procedure}
\end{figure}


%% file: content/preliminaries.tex
\section{Prelimiaries}
\label{sec:preliminaries}

We consider a training set $D = \left\{ (\xb_i, y_i) \right\}^{N}_{i=1} $, drawn \iid\ from the joint data distribution $\Pcal_{\Xcal \Ycal}$, where $\Xcal$ denotes the input space and $\Ycal \in \left\{ 1,2, \dots \Ccal \right \}$ denotes the label space with $\Ccal$ the number of classes. Let $\Probability_\text{id}$ be the marginal distribution on $\Xcal$, which is also referred to as the \textit{in-distribution}.

\subsection{Out-of-Distribution Detection}
\label{sec:ood}
\acrlong{ood} detection can be conceptualized as a task of distinguishing between two classes during inference, \emph{i.e.}, the objective of \gls{ood} detection is to determine whether a sample $x \in \Xcal$
originates from $\Probability_\text{id}$ or not. 
This task can be addressed by integrating a novelty detection mechanism during deployment that detects samples deviating from the known distribution, such as samples from an unrelated distribution with a label set disjoint from the in-distribution $\mathcal{Y}_{\text{id}}$, which should not be predicted by the model.
Formally, let $D_{ood}^{test}$ represent an out-of-distribution test set where the label space $\Ycal_\text{ood} \cap \Ycal_\text{id} = \emptyset$. The decision can be made employing level set estimation: $\Gcal_\lambda (x) = 1 \cbr{S(x) \geq \lambda}$, where samples with higher scores $S(x)$ are classified as \gls{id}, and vice versa. The threshold $\lambda$ is typically selected such that a substantial portion of in-distribution data \eg 95\% is accurately classified.

\subsection{Hyperbolic Space}
\label{sec:Hyperbolic_space}
Hyperbolic geometry is a non-Euclidean geometry characterized by constant negative curvature. 
It diverges from Euclidean geometry by permitting multiple parallel lines that pass through a point and are parallel to a given line. Various models, such as the Poincar\'e disk, Poincar\'e half-plane, Beltrami--Klein, and Lorentz (Hyperboloid/Minkowski) models, have been employed to depict Hyperbolic space \cite{mettesHyperbolicDeepLearning2023, leeRiemannianManifolds1997}. 
Although the Poincar\'e disk has been widely used in the \gls{dl} community \cite{vanspenglerPoincarResNet2023, nickelPoincarEmbeddingsLearning2017, tifreaPoincarGloVeHyperbolic2018}, this study adopts the Lorentz model due to its simpler closed-form representation of geodesics \cite{lawLorentzianDistanceLearning2019, desaiHyperbolicImagetextRepresentations2024} and its robustness to numerical instabilities \cite{mishneNumericalStabilityHyperbolic2023}.
Consider the $(n+1)$-dimensional Minkowski space, represented by the vectors $\zb = \sbr{z_{\text{time}}, \zb_{\text{space}}}$ where $z_{\text{time}}\in\R$, $\zb_{\text{space}}\in\R^n$, $\zb\in\R^{n+1}$, equipped with the Lorentz product
\begin{equation}
    \inner\zb{\zb'}_\Lorentz = \inner{\zb_{\text{space}}}{\zb'_{\text{space}}} - z_{\text{time}}z'_{\text{time}}.
\end{equation}
The Lorentz model $\Lorentz^n_c$ of an $n$-dimensional Hyperbolic space with curvature $c$ is the manifold defined by $\inner\zb{\zb}_\Lorentz=-1/c$ with $z_{\text{time}}>0$, which gives $z_{\text{time}} = \sqrt{1/c + ||\zb_\text{space}||^2}$.

\begin{definition}[Geodesic]
A geodesic is the shortest path between two points on a curved surface or manifold, analogous to a straight line in Euclidean space. The \emph{Lorentz distance} between two points $\xb$, $\yb \in \Lorentz_c^n$ is
\begin{equation}
    d_\Lorentz(\xb, \yb) = \frac{1}{\sqrt{c}} \cosh ^ {-1} ( -c \inner\xb{\yb}_\Lorentz ).
\label{eq:lorentz_distance}
\end{equation}
\end{definition}

\begin{definition}[Tangent space]
The tangent space at the point $\zb \in \Lorentz_c^n$ is a Euclidean space of vectors orthogonal to $\zb$ according to the Lorentz product:
\begin{equation}
   \Tcal_\zb \Lorentz_c^n = \cbr{\vb \in \R^{n+1} : \inner{\zb}{\vb}_\Lorentz = 0}.
\end{equation}
Any vector $\ub \in \R^{n+1}$ can be projected to the tangent space $\Tcal_\zb \in \Lorentz_c^n$ via orthogonal projection:
\begin{equation}
   \vb =\mathrm{proj}_\zb (\ub) = \ub + c\,\zb \inner{\zb}{\ub}_\Lorentz.
\end{equation}
\end{definition}

\begin{definition}[Exponential and Logarithmic maps]
The \textit{exponential map} provides a way to map a vector from the tangent spaces onto the manifold. Formally, for a point $\zb$ on the hyperboloid, it is defined as
$\mathrm{expm}_\zb: \Tcal_\zb \Lorentz^n_c \rightarrow \Lorentz^n_c$ with the expression:
\begin{equation}
\xb = \mathrm{expm}_\zb (\vb) = \cosh (\sqrt{c} ||\vb||_\Lorentz) \zb + \frac{\sinh(\sqrt{c} ||\vb||_\Lorentz)}{\sqrt{c} ||\vb||_\Lorentz} \vb.
\label{eq:exponential_map}
\end{equation}
The inverse operation is the \textit{logarithmic map} $\mathrm{logm}_\zb: \Lorentz^n_c \rightarrow \Tcal_\zb \Lorentz^n_c$ that maps $\xb$ from the Lorentz manifold $\Lorentz^n_c$ back to $\vb$ in the tangent space:
\begin{equation}
\vb = \mathrm{logm}_\zb(\xb) = \frac{\cosh ^{-1} (-c \inner\zb{\xb}_\Lorentz)}{\sqrt{(c \inner\zb{\xb}_\Lorentz)^2 -1}}.
\label{eq:logarithmic_map}
\end{equation}
\end{definition}

For our approach we will consider these maps where in $\zb$ is the origin of the hyperboloid $(\Ob = [1/\sqrt{c}, \textbf{0}])$.

%% file: content/methodology.tex
\section{Hyperbolic Space for Out-of-Distribution Detection}
\label{sec:how_to_use_Hyperbolic_for_ood}

\subsection{Compact Representation of In-Distribution Embeddings}
\label{sec:learn_embeddings}

Our objective is to leverage the properties of Hyperbolic space to improve the effectiveness of learned embeddings in distinguishing between \gls{id} and \gls{ood} data.
To achieve this, we use a feature encoder $f_\theta: \Xcal \rightarrow \R^e$ that transforms the augmented input $\xb \in \Xcal$ into a high-dimensional image embedding $f_\theta(\xb)$, often referred to as penultimate layer features.

Since these embeddings reside in Euclidean space, we need to project them onto the hyperboloid $\Lorentz^n_c$ to operate within the Hyperbolic space.
We accomplish this by applying the exponential map, as defined in \cref{eq:exponential_map}. Following \cite{desaiHyperbolicImagetextRepresentations2024}, we focus solely on the \textit{space} component of the \textit{exponential map} to reduce computational complexity and eliminate the orthogonal projection.

\begin{wrapfigure}{r}{0.47\textwidth}
  \begin{center}
  \vspace{-3.5em}
   {\includegraphics[width=\linewidth,height=\linewidth]{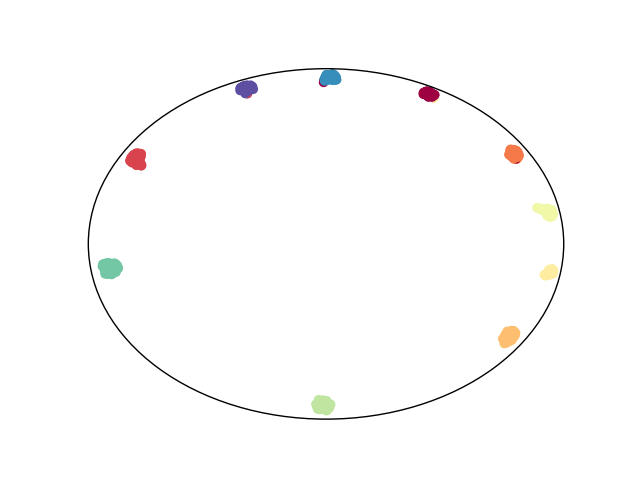}}
  \end{center}
  \vspace{-2em}
  \caption{UMAP visualization of learned feature embeddings of CIFAR-10 projected to the Poincar\'e disk using $\Lcal_{\text{hsup}}$.}
  \label{fig:poincare_embeddings}
\end{wrapfigure}

Once the embeddings are lifted into the Hyperbolic space, our next step involves optimizing them to encourage 
embeddings belonging to the same class to be close together, while simultaneously keeping apart clusters of samples from different classes. 
This optimization strategy follows a supervised contrastive learning approach  \cite{khoslaSupervisedContrastiveLearning2020} promoting low intra-class variation and high inter-class separation, but rather than relying on cosine similarity, we use the negative Lorentzian distance \cref{eq:lorentz_distance} as a similarity measure. 
The process solely involves training the parameters $\theta$ of the image feature encoder and the projection layer. It is guided by the loss function:

\begin{equation}
\Lcal_{\text{hsup}} = \sum_{i \in I} \frac{-1}{|P(i)|} \sum_{p \in P(i)} \log \frac{\exp \rbr{-d_\Lorentz (\zb_i, \zb_p)/\tau}}
{\sum_{a \in A(i)} \exp \rbr{-d_\Lorentz \rbr{\zb_i, \zb_a}/\tau}}.
\label{eq:Hyperbolic_contrastive_loss}
\end{equation}
Here, $P(i) = \cbr{p \in A(i) : \tilde{\yb}_p = \tilde{\yb}_i}$ is the set of indices of all positives in the multiviewed batch distinct from $i$, and $|P(i)|$ is its cardinality. 


As illustrated in \cref{fig:poincare_embeddings}, this function effectively improves the formation of tight clusters on the manifold while compelling in-distribution embeddings move towards the boundary. This characteristic, also observed in prior research \cite{khrulkovHyperbolicImageEmbeddings2020, ghadimiatighHyperbolicImageSegmentation2022}, demonstrates that broader concepts in the hierarchy tend to cluster closer to the origin, while finer, more specific concepts are pushed towards the boundary. Leveraging this feature proves advantageous for both the classification of \gls{id} embeddings (discussed in \cref{sec:classification_id}) and the generation of new embeddings (explored in \cref{sec:synthesis_outliers}).

\subsection{Classification of In-Distribution Embeddings}
\label{sec:classification_id}
Traditional multi-class classification often relies on methods like \gls{mlr}. However, in our Hyperbolic setup, constructing a classifier involves determining the distance to margin hyperplanes within the Lorentz model. In this context, for a point $\pb \in \Lorentz^n_c$ and a vector $\wb \in \Tcal_\pb \Lorentz^n_c$, the hyperplane passing through $\pb$ and perpendicular to $\wb$ is represented by:
\begin{equation}
    H_{\wb,\pb} = \cbr{\zb \in \Lorentz^n_c | \inner{\wb}{\zb}_\Lorentz = 0}
\end{equation}

To address the non-convex optimization condition $\inner{\wb}{\wb}_\Lorentz>0$ we follow Bdeir \etal \cite{bdeir2024fully}.
Define the Lorentz hyperplane as
\begin{equation}
   \tilde{H}_{\wb} = \cbr{\zb \in \Lorentz^n_c | \inner{\wb}{\zb}_\Lorentz = 0},
   \label{eq:hyperplane_lorentz}
\end{equation}
and rewrite $\wb$ using its distance to the origin $a$ and its orientation $\ob$,
\begin{equation}
    w_{\text{time}} = \sinh(a\sqrt{c})||\ob||, \qquad \wb_{\text{space}} = \cosh(a\sqrt{c}) \ob.
\end{equation}
Using the distance to the hyperplane,
\begin{equation}
    d_\Lorentz(\zb, \tilde{H}_{\wb}) = \frac{1}{\sqrt{c}} \left|
    \sinh^{-1} \rbr{\sqrt{c}\frac{\inner{\wb}{\zb}_\Lorentz}{||\wb||_\Lorentz}}
    \right|,
\end{equation}
the model’s logit for class $k$ is given by
\begin{equation}
    \ell_{\wb_k}(\zb) = \mathrm{sign}(\inner{\wb_k}{\zb}_\Lorentz) ||\wb_k||_\Lorentz d_\Lorentz(\zb, \tilde{H}_{\wb_k}).
\end{equation}

Leveraging the inherent clustering of similar embeddings due to the Hyperbolic contrastive loss \cref{eq:Hyperbolic_contrastive_loss}, the model therefore constructs an effective hyperplane for robust class discrimination.


\subsection{Regularization with Synthetic Hyperbolic Outliers}
\label{sec:synthesis_outliers}
The absence of explicit knowledge regarding unknown inputs during training leads to challenges. Models, usually optimized solely on \gls{id} data, may produce decision boundaries beneficial for tasks like classification but unsuitable for \gls{ood} detection.
Incorporating external information from these unknowns through $D_{\text{ood}}^{\text{aux}}$ during training helps in assigning lower scores to randomly selected outliers, thereby establishing a clear boundary between \gls{id} and \gls{ood} samples. These chosen outliers should not only be diverse but also informative, resembling \gls{id} samples closely.

Building upon previous research that utilized synthetic outliers sampled in the feature space \cite{duVOSLearningWhat2022, taoNonparametricOutlierSynthesis2022}, we propose a regularization approach for the model employing the Hyperbolic space, \ie Hyperbolic synthetic outliers. Our central concept leverages the observation that broader concepts in the visual hierarchy are closer to the origin, allowing us to sample from them and enhance the compactness of the \gls{id} representation.

Specifically, let $\mathcal{Z} = \{\mathbf{z}_1, \mathbf{z}_2, \dots, \mathbf{z}_n\}$ denote the set of \gls{id} embeddings from the training data, where $\mathbf{z}_i \in \mathbb{L}^n_c$ represents a feature in the Hyperbolic space. For any embedding $\mathbf{z}_i \in \mathcal{Z}$, we compute its $L_2$ norm with respect to the origin. If an embedding has a small $L_2$ norm, it is likely to be uncertain. Therefore, we select embeddings with the smallest $L_2$ distances. Once we have identified a set of ambiguous \gls{id} embeddings, we synthesize outliers by sampling from a Hyperbolic wrapped Gaussian distribution \cite{naganoWrappedNormalDistribution2019} centered around the selected uncertain \gls{id} embedding $\sbb \sim \Wcal\Gcal \rbr{\zb_i, \sigma^2 \Ib}$ where the $\sbb \in \Lorentz^n_c$ denotes the synthesized outlier around $\zb_i$, and $\sigma^2$ modulates the variance. For each uncertain \gls{id} embedding we produce a set of synthetic outliers $\Scal = \cbr{\sbb_1, \sbb_2, \dots, \sbb_n}$. 

Then, we filter the synthetic outliers $\mathbf{\sbb}_j$ by selecting those with the highest $L_2$ distances (indicating higher certainty), as long as they are less than the \gls{id} embedding $\zb_i$ from which they originated $||\sbb_j - \zb_i||_2 < ||\zb_i||_2$. This procedure is visualized in \cref{fig:sampling-procedure}.

The final collection $\mathcal{V}$ of accepted synthetic outliers will be used for the binary training objective:
\begin{equation}
    \Lcal_{\text{uncertainty}} = \E_{\vb \sim \Vcal} \sbr{-\log \frac{1}{1+ e^{\ell_\wb(\vb)}}} + \E_{\xb\sim \Probability_{\text{in}}} \sbr{-\log \frac{e^{\ell_\wb(\zb)}}{1+ e^{{\ell_\wb(\zb)}}}}.
    \label{eq:uncertainty_loss}
\end{equation}
The loss function takes both the \gls{id} and synthesized outlier embeddings and aims to estimate a hyperplane ${H}_{\ob,a}(\cdot)$ that divides them through the binary cross-entropy loss. This process pushes the synthetic outliers embeddings towards the origin while maintaining the \gls{id} samples near the boundary. Consequently, real \gls{ood} data tends to be positioned further from the boundary, facilitating easier identification. The final objective is formulated as a combination of the metric loss in \cref{eq:Hyperbolic_contrastive_loss} and the regularization loss \cref{eq:uncertainty_loss} with a weighting factor $\alpha$:
\begin{equation}
    \Lcal = \Lcal_{\text{hsup}} + \alpha \Lcal_{\text{uncertainty}}.
\end{equation}
By varying $\alpha$, the relative importance of the synthetic outliers can be controlled, enabling flexibility in the optimization process.

%% file: content/experiments.tex
\section{Experimental Results}
\label{sec:experiments_results}

\subsection{Setup}
\label{sec:setup}

\paragraph{\textbf{Datasets.}} Following common benchmarks in the literature, we consider CIFAR-10 \cite{krizhevskyLearningMultipleLayers}, CIFAR-100 \cite{krizhevskyLearningMultipleLayers}, and ImageNet-200 \cite{dengImagenetLargescaleHierarchical2009,hendrycks2021many} as our in-distribution dataset $D_{\text{id}}$. We refrain from tuning hyperparameters specifically for each $D_{\text{ood}}^{\text{test}}$ dataset, thereby maintaining its unknown nature, akin to real-world scenarios. Subsequently, our considered validation out-of-distribution $D_{\text{ood}}^{\text{val}}$ datasets include LSUN \cite{yuLsunConstructionLargescale2015}, LSUN Resize \cite{yuLsunConstructionLargescale2015}, and iSUN \cite{xu2015turkergaze}. For $D_{\text{ood}}^{\text{test}}$, we employ a suite of natural image datasets following the OpenOOD framework \cite{yang2022openood,zhangOpenOODV1Enhanced2023}, which comprises SVHN \cite{netzerReadingDigitsNatural2011},  Places365 \cite{zhouPlaces10Million2017}, Textures \cite{cimpoiDescribingTexturesWild2014}, and MNIST \cite{dengMNISTDatabaseHandwritten2012} for experiments involving CIFAR-10 and CIFAR-100. For experiments involving ImageNet-200, we utilize SSB \cite{vaze2022openset}, Ninco \cite{bitterwolf2023fixing} as Near-OOD datasets, and iNaturalist \cite{vanhornInaturalistSpeciesClassification2018a}, Textures \cite{cimpoiDescribingTexturesWild2014} and OpenImage-O \cite{wangViMOutofdistributionVirtuallogit2022} as Far-OOD datasets.


\paragraph{\textbf{Out-of-Distribution detection scores.}} 
 \gls{ood} scores are determined using a distance-based approach during testing. An input $\zb \in \Lorentz_c^n$ is classified as \gls{ood} if it exhibits significant distance from the \gls{id} data within the Hyperbolic embedding space. Our default method is a straightforward \gls{knn} in the Hyperbolic space, which avoids imposing any distributional assumptions on the space and uses the potential of the exponential growth in the Hyperbolic space. Note that the distance for \gls{knn} is the negative Lorentzian distance \cref{eq:lorentz_distance}. We tune the value of $k$ neighbours using the validation set $D_{\text{ood}}^{\text{val}}$. The parameter that yields the best \gls{auroc} is used for the final test in $D_{\text{ood}}^{\text{test}}$.

\paragraph{\textbf{Training details.}} 
To enable fair comparison with other works, we adopted the OpenOOD \cite{yang2022openood,zhangOpenOODV1Enhanced2023} benchmark setup which uses ResNet-18 for CIFAR-10/100 and ImageNet-200.
HOD comprises two main components: (1) mapping the embeddings from the final layer to hyperbolic space, and (2) training these embeddings using the contrastive loss, as detailed in \cref{sec:learn_embeddings}. The method then employs \gls{knn} to identify \gls{ood} samples. The incorporation of synthetic outliers in the HOD strategy is considered when the hyperparameter $\alpha > 0$.
As we obtain features in the hyperbolic space with changes only in the last layer, our method applies to deeper encoders, and we ablate the backbone depth in the supplementary material. We set the embedding dimension to 128 for the Hyperbolic projection head. To optimize the network, we employ AdamW \cite{loshchilov2017decoupled} with a weight decay of $0.2$ and $(\beta_1, \beta_2) = (0.9, 0.98)$. Weight decay is disabled for gains, biases, and learnable scalars \ie curvature $c$. The training process spans $20K$ iterations with a batch size of $256$ for {CIFAR} datasets and $512$ for ImageNet-200. The maximum learning rate initiates at $1\times10^{-3}$, linearly increasing for the initial $400$ iterations, followed by cosine decay to zero \cite{loshchilov2016sgdr}. We set the weight of synthetic outliers as $\alpha=0.1$ and commence sampling after $1,000$ iterations.

\paragraph{\textbf{Evaluation metrics.}} We report the following metrics: \textbf{(1)} the false positive rate \gls{fpr95} of \gls{ood} samples when the true positive rate of \gls{id} samples is at 95\% and \textbf{(2)} the \gls{auroc}. Our evaluation consists of 3 independent training runs for all $D_{\text{id}}$. We average scores for the same random seed across datasets, and report the mean and standard deviation for the seed.
We further evaluate the statistical significance of observed differences by comparing our method to the best-performing baseline on each dataset and metric using a $t$-test, assuming statistical significance for $p$-values below $0.05$ and denote this in \textbf{Bold}.

\subsection{Evaluation on Hyperbolic Out-of-Distribution Scores}
\label{sec:evaluation_hyperbolic_ood_scores}
\begin{table}[t!]
\caption{\gls{ood} detection performance on {CIFAR-10} as  $D_{\text{id}}$. The values are the mean and standard deviation after training each method three times on ResNet-18. \textbf{Bold} if there is a statistical significance difference.}
  \label{tab:evaluation_ood_scores_cifar10}%
  \resizebox{\textwidth}{!}{
  \begin{tabular}{rcccccccccc}
   \toprule
    \multirow{3}[4]{*}{Methods} & \multicolumn{9}{c}{$D^{\text{val}}_{\text{out}}$}                                                              \\
    \cmidrule{2-7}
          & \multicolumn{2}{c}{LSUN} & \multicolumn{2}{c}{LSUN Resize} & \multicolumn{2}{c}{iSUN} & \multicolumn{2}{c}{Average} &  \\
\cmidrule{2-7}          & \gls{fpr95}$\downarrow$ & \gls{auroc}$\uparrow$ & \gls{fpr95}$\downarrow$ & \gls{auroc}$\uparrow$ & \gls{fpr95}$\downarrow$ & \gls{auroc}$\uparrow$ & \gls{fpr95}$\downarrow$ & \gls{auroc}$\uparrow$    \\
    \midrule
    EBO & $39.1_{\pm 1.8}$ & $95.2_{\pm 0.2}$ & $29.6_{\pm 6.1}$ & $95.0_{\pm 1.1}$ & $28.6_{\pm 6.1}$ & $95.3_{\pm 1.1}$ & $32.4_{\pm 5.4}$ & $95.2_{\pm 0.1}$\\
    Softmax & $17.6_{\pm 1.8}$ & $96.9_{\pm 0.2}$ & $31.6_{\pm 7.0}$ & $94.9_{\pm 1.2}$ & $30.3_{\pm 7.1}$ & $95.1_{\pm 1.2}$ & $26.5_{\pm 7.7}$ & $95.6_{\pm 1.1}$ \\
    Distance Origin & $57.2_{\pm 3.9}$ & $90.0_{\pm 1.3}$ & $86.9_{\pm 7.6}$ & $79.3_{\pm 4.5}$ & $86.0_{\pm 8.6}$ & $79.4_{\pm 4.9}$ & $76.7_{\pm 16.9}$ & $82.9_{\pm 6.1}$ \\
    $\Lcal_{\text{hsup}}$ + EBO & $32.4_{\pm 5.6}$ & $95.2_{\pm 0.9}$ & $57.7_{\pm 0.4}$ & $88.1_{\pm 0.2}$ & $61.8_{\pm 2.8}$ & $87.4_{\pm 1.0}$ & $50.6_{\pm 15.9}$ & $90.2_{\pm 4.3}$ \\
    $\Lcal_{\text{hsup}}$ + Softmax & $9.3_{\pm 0.5}$ & $97.9_{\pm 0.1}$ & $41.6_{\pm 0.5}$ & $90.8_{\pm 2.5}$ & $46.0_{\pm 0.7}$ & $91.5_{\pm 0.0}$ & $32.3_{\pm 20.0}$ & $93.4_{\pm 3.9}$ \\
    $\Lcal_{\text{hsup}}$ + Distance Origin & $10.3_{\pm 0.1}$ & $97.6_{\pm 0.1}$ & $62.0_{\pm 0.8}$ & $87.3_{\pm 0.1}$ & $66.1_{\pm 1.1}$ & $86.5_{\pm 0.2}$ & $46.1_{\pm 31.1}$ & $90.5_{\pm 6.2}$ \\
    \midrule
    HOD ($\alpha=0$) & $9.9_{\pm 2.7}$ & $98.4_{\pm 0.4}$ & $\mathbf{8.9_{\pm 5.8}}$ & $\mathbf{98.8_{\pm 0.4}}$ & $\mathbf{5.9_{\pm 5.1}}$ & $\mathbf{98.7_{\pm 0.6}}$ & $\mathbf{8.2_{\pm 2.1}}$ & $\mathbf{98.6_{\pm 0.2}}$ \\
    HOD ($\alpha=0.1$) & $10.3_{\pm 0.6}$ & $97.2_{\pm 0.6}$ & $\mathbf{2.7_{\pm 1.7}}$ & $\mathbf{99.3_{\pm 0.2}}$ & $\mathbf{1.8_{\pm 0.2}}$ & $\mathbf{99.4_{\pm 0.0}}$ & $\mathbf{4.9_{\pm 4.6}}$ & $\mathbf{98.6_{\pm 1.2}}$\\
    \bottomrule
    \end{tabular}%
    }
      
\end{table}


This section evaluates the effectiveness of different \gls{ood} detection scores. We compare \gls{knn} with three established methods previously used in the Hyperbolic space: Energy-score (EBO) \cite{vanspenglerPoincarResNet2023}, Softmax score \cite{guoClippedHyperbolicClassifiers2022} and the Distance to the origin \cite{khrulkovHyperbolicImageEmbeddings2020, ghadimiatighHyperbolicImageSegmentation2022}.
To select the best \gls{ood} detection strategy, we evaluate the models using the $D_\text{out}^{\text{val}}$ metric and keep the $D_\text{out}^{\text{test}}$ unseen. We follow the procedure outlined in \cref{sec:setup} but with a cross-entropy loss as the training objective. Subsequently, we evaluate the proposed $\Lcal_{\text{hsup}}$ \cref{eq:Hyperbolic_contrastive_loss} by pre-training the model and fine-tuning only the classification head.

The results, presented in \cref{tab:evaluation_ood_scores_cifar10}, demonstrate that our detection score based on the Hyperbolic \gls{knn} outperforms the baselines both with and without synthetic outliers sampling. Interestingly, pre-training the network with the $\Lcal_{\text{hsup}}$ improves results only for distance-based methods.

\subsection{Main Results}
\label{sec:results}

Our comparison includes the competing approaches EBO~\cite{liuEnergybasedOutofdistributionDetection2020}, ODIN~\cite{hsuGeneralizedOdinDetecting2020a}, MLS~\cite{hendrycksScalingOutofdistributionDetection2022}, GradNorm~\cite{huangImportanceGradientsDetecting2021a}, ViM~\cite{wangViMOutofdistributionVirtuallogit2022}, KNN~\cite{sunOutofdistributionDetectionDeep2022}, CIDER~\cite{mingHowExploitHyperspherical2023}, VOS~\cite{duVOSLearningWhat2022}, and NPOS~\cite{taoNonparametricOutlierSynthesis2022}.
In particular, CIDER is also based on non-Euclidean geometry, but it uses hyperspherical embeddings which do not feature exponential volume growth instead of hyperbolic ones. Furthermore, synthetic outliers are used by both VOS, which models \gls{id} embeddings as a Gaussian mixture and samples synthetic outliers from the low-likelihood region of the feature space, and NPOS, which takes a non-parametric approach similar to this work.

\begin{table}[t!]
\caption{\gls{ood} detection performance on CIFAR-10 as  $D_{\text{id}}$. The values are the mean and standard deviation after training each method three times on ResNet-18. \textbf{Bold} if there is a statistical significance difference.}
  \label{tab:train_from_scratch_results_c10}%
  \resizebox{\textwidth}{!}{
  \begin{tabular}{rccccccccccc}
   \toprule
    \multirow{3}[4]{*}{Methods} & \multicolumn{9}{c}{$D^{\text{test}}_{\text{out}}$}                                                              \\
    \cmidrule{2-9}
          & \multicolumn{2}{c}{Texture} & \multicolumn{2}{c}{Places365} & \multicolumn{2}{c}{SVHN} & \multicolumn{2}{c}{MNIST} & \multicolumn{2}{c}{Average} &  \\
\cmidrule{2-9}          & \gls{fpr95}$\downarrow$ & \gls{auroc}$\uparrow$ & \gls{fpr95}$\downarrow$ & \gls{auroc}$\uparrow$ & \gls{fpr95}$\downarrow$ & \gls{auroc}$\uparrow$ & \gls{fpr95}$\downarrow$ & \gls{auroc}$\uparrow$  & \gls{fpr95}$\downarrow$ & \gls{auroc}$\uparrow$  &  \\
    \midrule
    EBO \cite{liuEnergybasedOutofdistributionDetection2020}  & $51.8_{\pm 7.5}$  & $89.5_{\pm 0.9}$ & $54.5_{\pm 8.0}$ & $89.3_{\pm 1.0}$ & $35.1_{\pm 7.5}$ & $91.8_{\pm 1.2}$ & $25.0_{\pm 15.8}$ & $93.0_{\pm 2.6}$ & $41.7_{\pm 6.5}$ & $90.9_{\pm 1.0}$\\
    ODIN \cite{hsuGeneralizedOdinDetecting2020a}  & $67.7_{\pm 13.6}$ & $86.9_{\pm 2.8}$ & $70.3_{\pm 8.5}$ & $85.1_{\pm 1.5}$ & $68.6_{\pm 0.6}$ & $84.6_{\pm 1.0}$ & $23.8_{\pm 15.1}$ & $95.2_{\pm 2.4}$ & $57.6_{\pm 5.2}$ & $88.0_{\pm 0.8}$\\
    MLS  \cite{hendrycksScalingOutofdistributionDetection2022} & $51.7_{\pm 7.5}$  & $89.4_{\pm 0.9}$ & $54.8_{\pm 8.0}$ & $89.1_{\pm 0.9}$ & $35.1_{\pm 7.5}$ & $91.7_{\pm 1.2}$ & $25.2_{\pm 15.9}$ & $94.2_{\pm 3.0}$ & $41.7_{\pm 6.5}$ & $91.1_{\pm 1.1}$\\
    GradNorm \cite{huangImportanceGradientsDetecting2021a} & $98.1_{\pm{0.6}}$& $52.1_{\pm 5.0}$ & $92.5_{\pm 2.8}$ & $60.5_{\pm 6.5}$ & $91.7_{\pm 3.0}$ & $53.9_{\pm 7.8}$ & $85.4_{\pm 6.0}$  & $63.7_{\pm 9.0}$ & $91.9_{\pm 2.7}$  & $57.5_{\pm 3.9}$\\
    ViM \cite{wangViMOutofdistributionVirtuallogit2022}  & $21.1_{\pm 2.2}$  & $95.2_{\pm 0.4}$ & $41.4_{\pm 2.6}$ & $89.5_{\pm 0.5}$ & $19.3_{\pm 0.5}$ & $94.5_{\pm 0.6}$ & $19.3_{\pm 1.1}$  & $94.8_{\pm 0.5}$ & $25.3_{\pm 0.6}$ & $93.5_{\pm 0.3}$\\
    KNN \cite{sunOutofdistributionDetectionDeep2022}  & $24.1_{\pm 0.7}$  & $93.2_{\pm 0.3}$ & $30.4_{\pm 0.8}$ & $91.8_{\pm 0.3}$ & $22.6_{\pm 1.6}$ & $92.7_{\pm 0.4}$ & $20.1_{\pm 1.7}$  & $94.3_{\pm 0.5}$ & $24.3_{\pm 0.5}$  & $93.0_{\pm 0.2}$\\
    CIDER \cite{mingHowExploitHyperspherical2023} & $25.1_{\pm  3.3}$ &  $93.7_{\pm  0.4}$ &	  	  $25.0_{\pm  1.4}$ &               $93.8_{\pm  0.7}$ &     	$8.0 _{\pm  0.4}$ & $98.1_{\pm  0.1}$ &		$24.8_{\pm  2.8}$ & $93.3_{\pm  1.1}$ & $20.7_{\pm  0.9}$ & $94.7_{\pm  0.4}$ \\
    VOS \cite{duVOSLearningWhat2022}  & $48.6_{\pm 4.2}$ & $89.8_{\pm 0.8}$ & $56.4_{\pm 1.2}$ & $89.2_{\pm 0.3}$ & $32.2_{\pm 7.6}$ & $92.1_{\pm 1.6}$ & $20.0_{\pm 1.5}$ & $94.9_{\pm 0.4}$ & $39.3_{\pm 3.4}$ & $91.5_{\pm 0.7}$\\
    NPOS \cite{taoNonparametricOutlierSynthesis2022} & $26.8_{\pm 1.6}$  & $91.9_{\pm 1.0}$ & $31.2_{\pm 3.0}$ & $91.4_{\pm 0.7}$ & $8.2_{\pm 3.3}$  & $98.5_{\pm 0.5}$ & $23.7_{\pm 2.2}$  & $93.4_{\pm 1.6}$ & $22.5_{\pm 2.1}$ & $93.8_{\pm 0.8}$\\
    \midrule
    HOD ($\alpha=0$) & $19.3_{\pm 2.0}$ & $\mathbf{97.8_{\pm 0.1}}$ & $\mathbf{10.6_{\pm 0.3}}$ & $\mathbf{98.4_{\pm 0.4}}$ & $12.4_{\pm 0.4}$ & $98.0_{\pm 0.4}$ & $21.1_{\pm 6.3}$ & $96.8_{\pm 1.5}$ & $\mathbf{15.8_{\pm 1.2}}$ & $\mathbf{97.8_{\pm 0.4}}$ \\
    HOD ($\alpha=0.1$) & $\mathbf{16.1_{\pm 0.6}}$ & $97.3_{\pm 0.6}$ & $\mathbf{10.3_{\pm 0.2}}$ & $\mathbf{98.7_{\pm 0.4}}$ & $16.2_{\pm 0.2}$ & $93.2_{\pm 0.6}$ & $19.7_{\pm 0.2}$ & $94.0_{\pm 0.8}$ & $\mathbf{15.6_{\pm 0.2}}$ & $95.3_{\pm 0.3}$ \\
    \bottomrule
    \end{tabular}%
    }
\end{table}

\begin{table}[!t]
\caption{\gls{ood} detection performance on CIFAR-100 as  $D_{\text{id}}$. The values are the mean and standard deviation after training each method three times on ResNet-18. \textbf{Bold} if there is statistical significance difference.}
  \label{tab:train_from_scratch_results_c100}%
  \resizebox{\textwidth}{!}{
  \begin{tabular}{rccccccccccc}
   \toprule
    \multirow{3}[4]{*}{Methods} & \multicolumn{9}{c}{$D^{\text{test}}_{\text{out}}$}                                                              \\
    \cmidrule{2-9}
          & \multicolumn{2}{c}{Texture} & \multicolumn{2}{c}{Places365} & \multicolumn{2}{c}{SVHN} & \multicolumn{2}{c}{MNIST} & \multicolumn{2}{c}{Average} &  \\
\cmidrule{2-9}          & \gls{fpr95}$\downarrow$ & \gls{auroc}$\uparrow$ & \gls{fpr95}$\downarrow$ & \gls{auroc}$\uparrow$ & \gls{fpr95}$\downarrow$ & \gls{auroc}$\uparrow$ & \gls{fpr95}$\downarrow$ & \gls{auroc}$\uparrow$  & \gls{fpr95}$\downarrow$ & \gls{auroc}$\uparrow$  &  \\
    \midrule
    EBO \cite{liuEnergybasedOutofdistributionDetection2020}  & $62.4_{\pm 2.5}$ & $78.4_{\pm 1.0}$ & $58.7_{\pm 1.0}$ & $79.2_{\pm 0.3}$ & $53.8_{\pm 3.8}$ & $82.0_{\pm 2.2}$ & $53.7_{\pm 5.4}$ & $79.6_{\pm 1.4}$ & $57.1_{\pm 2.5}$ & $79.8_{\pm 0.8}$\\
    ODIN \cite{hsuGeneralizedOdinDetecting2020a} & $62.4_{\pm 3.6}$ & $79.3_{\pm 1.3}$ & $60.5_{\pm 0.7}$ & $79.1_{\pm 0.3}$ & $67.4_{\pm 4.8}$ & $74.7_{\pm 0.9}$ & $44.6_{\pm 3.4}$ & $84.4_{\pm 1.4}$ & $58.7_{\pm 1.0}$ & $79.4_{\pm 0.3}$\\
    MLS \cite{hendrycksScalingOutofdistributionDetection2022}  & $62.4_{\pm 2.7}$ & $78.4_{\pm 1.0}$ & $58.8_{\pm1.1}$ & $79.4_{\pm 0.3}$ & $53.8_{\pm 4.0}$ & $81.9_{\pm 1.9}$ & $51.5_{\pm 3.9}$ & $79.6_{\pm 1.6}$ & $56.6_{\pm 1.6}$ & $79.8_{\pm 0.7}$\\
    GradNorm \cite{huangImportanceGradientsDetecting2021a} & $92.4_{\pm 0.7}$ & $64.6_{\pm 0.2}$ & $84.6_{\pm 1.0}$ & $69.5_{\pm 1.0}$ & $69.2_{\pm 10.6}$ & $73.6_{\pm 9.2}$ & $85.6_{\pm 1.9}$ & $66.1_{\pm 1.2}$ & $82.9_{\pm 2.7}$ & $68.4_{\pm 2.3}$\\
    ViM  \cite{wangViMOutofdistributionVirtuallogit2022} & $46.9_{\pm 2.8}$ & $85.9_{\pm 1.0}$ & $58.4_{\pm 5.8}$ & $77.4_{\pm 3.0}$ & $46.6_{\pm 6.6}$ & $82.9_{\pm 4.5}$ & $47.1_{\pm 1.8}$ & $82.2_{\pm 1.5}$ & $49.8_{\pm 0.7}$ & $82.1_{\pm 0.2}$\\
    KNN \cite{sunOutofdistributionDetectionDeep2022}  & $57.0_{\pm 5.1}$ & $81.8_{\pm 3.2}$ & $58.7_{\pm 5.9}$ & $80.7_{\pm 2.8}$ & $51.9_{\pm 3.7}$ & $84.0_{\pm 1.3}$ & $47.4_{\pm 5.6}$ & $83.0_{\pm 1.7}$ & $53.7_{\pm 0.4}$ & $82.4_{\pm 0.2}$\\
    CIDER \cite{mingHowExploitHyperspherical2023} & $54.4_{\pm 2.6}$ & $82.2_{\pm 1.9}$ &	   	     $69.3_{\pm 1.8}$ & $74.4_{\pm 0.6}$ & $17.8_{\pm 2.8}$ & $97.2_{\pm 0.3}$ & $75.3_{\pm 4.2}$ & $68.1_{\pm 4.0}$ & $54.2_{\pm 1.2}$ & $80.5_{\pm 0.7}$ \\
    VOS \cite{duVOSLearningWhat2022}  & $63.7_{\pm 2.0}$ & $78.3_{\pm 0.5}$ & $57.8_{\pm 0.2}$ & $79.7_{\pm 0.2}$ & $44.8_{\pm 5.6}$ & $84.6_{\pm 3.2}$ & $43.9_{\pm 1.1}$ & $81.5_{\pm 1.0}$ & $52.5_{\pm 0.7}$ & $81.0_{\pm 0.4}$ \\
    NPOS \cite{taoNonparametricOutlierSynthesis2022} & $46.4_{\pm 0.3}$ & $86.0_{\pm 0.9}$ & $58.3_{\pm 0.4}$ & $78.8_{\pm 0.7}$ & $30.3_{\pm 1.0}$ & $92.3_{\pm 0.8}$ & $64.2_{\pm 7.1}$ & $76.6_{\pm 5.1}$ & $49.8_{\pm 1.9}$ & $83.4_{\pm 1.6}$\\
    \midrule
    HOD ($\alpha=0$) & $\mathbf{34.1_{\pm 2.5}}$ & $\mathbf{89.0_{\pm 1.6}}$ & $\mathbf{36.6_{\pm 2.4}}$ & $84.6_{\pm 1.0}$ & $28.5_{\pm 2.2}$ & $90.5_{\pm 0.2}$ & $\mathbf{14.9_{\pm 1.0}}$ & $\mathbf{93.7_{\pm 0.2}}$ & $\mathbf{28.5_{\pm 0.4}}$ & $\mathbf{89.4_{\pm 0.5}}$\\
    HOD ($\alpha=0.1$) & $\mathbf{32.4_{\pm 1.9}}$ & $\mathbf{89.3_{\pm 0.7}}$ & $\mathbf{37.6_{\pm 1.5}}$ & $83.5_{\pm 0.5}$ & $31.6_{\pm 1.7}$ & $88.8_{\pm 0.5}$ & $\mathbf{14.4_{\pm 1.7}}$ & $\mathbf{94.0_{\pm 0.8}}$ & $\mathbf{29.0_{\pm 0.5}}$ & $\mathbf{88.9_{\pm 1.1}}$\\
    \bottomrule
    \end{tabular}%
    }
\end{table}

\begin{figure}
    \centering
    \includegraphics{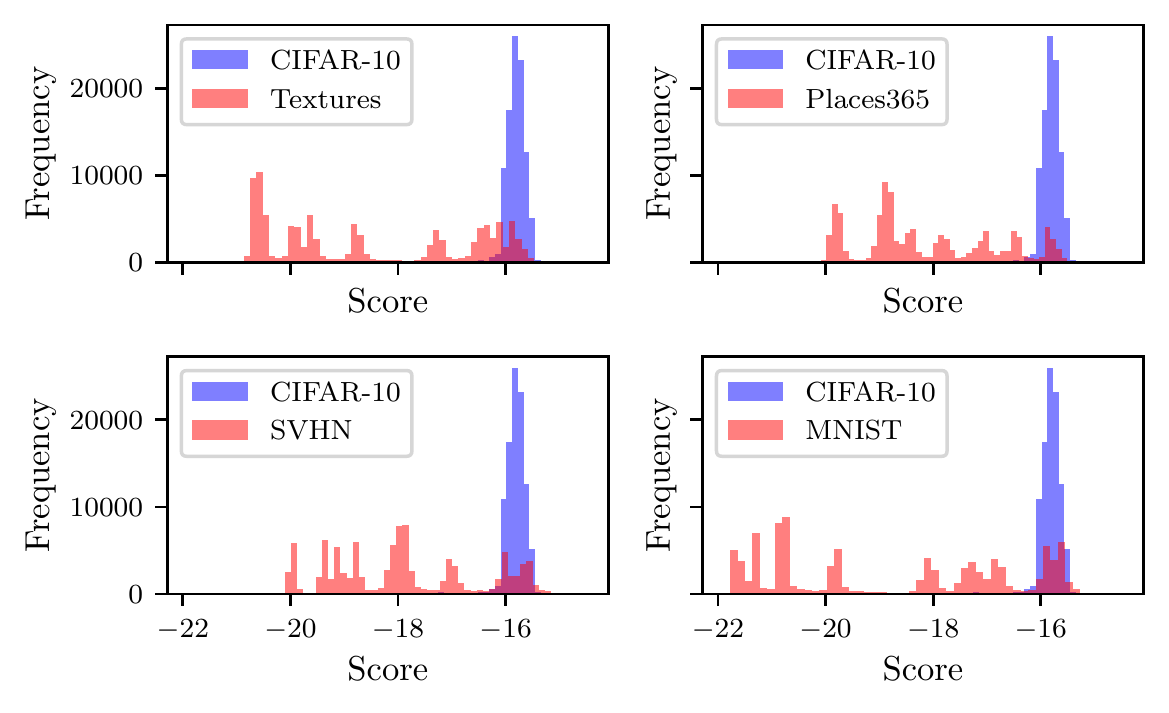}
    \caption{Histogram displaying scores for both \gls{id} and \gls{ood} datasets. Visualization reveals that distances to the in-distribution data are consistently lower than those to the out-of-distribution data. 
    }
    \label{fig:histogram_scores_cifar10}
\end{figure}

\paragraph{\textbf{HOD outperforms Euclidean space methods.}} Our evaluation demonstrates superiority of the Hyperbolic space over traditional Euclidean methods in \gls{ood} detection tasks. On {CIFAR-10}, HOD reduces the \gls{fpr95} from 20.7\% using CIDER to 15.6\% as shown in \cref{tab:train_from_scratch_results_c10}. On {CIFAR-100}, the reduction is even more substantial, decreasing from 49.8\% using NPOS or ViM to 28.5\% as shown in \cref{tab:train_from_scratch_results_c100}. Notably, HOD surpasses all competing methods by a substantial margin, as highlighted by statistically significant differences.
\Cref{fig:histogram_scores_cifar10} compares the Hyperbolic \gls{knn} scores between \gls{id} and \gls{ood} test data. The \gls{id} scores cluster significantly higher, indicating closer distances to the reference points compared to the \gls{ood} distribution. This clear distinction in Hyperbolic \gls{knn} scores makes them effective for differentiating \gls{id} and \gls{ood} data, leading to more efficient \gls{ood} detection.

The results on the ImageNet-200 dataset, reported in \cref{tab:train_from_scratch_results_imagenet200}, are more nuanced. While the margin between HOD and Euclidean approaches is not as large, the former still improves \gls{fpr95} in both Near-OOD datasets {SBB} and {Ninco}, and for two of the Far-OOD datasets, especially {OpenImages-O}. This suggests that the advantage of Hyperbolic space might be dataset-dependent and potentially more pronounced in scenarios with fewer classes.

\begin{table}[!t]
\caption{\gls{ood} detection performance on ImageNet-200 as  $D_{\text{id}}$. The values are the mean and standard deviation after training each method three times on ResNet-18. \textbf{Bold} if there is statistical significance difference.}
  \label{tab:train_from_scratch_results_imagenet200}%
  \resizebox{\textwidth}{!}{
  \begin{tabular}{rccccccccccccc}
   \toprule
    \multirow{3}[4]{*}{Methods} & \multicolumn{11}{c}{$D^{\text{test}}_{\text{out}}$}                                                              \\
    \cmidrule{2-11}
          & \multicolumn{2}{c}{SSB} & \multicolumn{2}{c}{Ninco} & \multicolumn{2}{c}{iNaturalist} & \multicolumn{2}{c}{Texture} & \multicolumn{2}{c}{OpenImage-O} & \multicolumn{2}{c}{Average} &  \\
\cmidrule{2-11}          & \gls{fpr95}$\downarrow$ & \gls{auroc}$\uparrow$ & \gls{fpr95}$\downarrow$ & \gls{auroc}$\uparrow$ & \gls{fpr95}$\downarrow$ & \gls{auroc}$\uparrow$ & \gls{fpr95}$\downarrow$ & \gls{auroc}$\uparrow$ & \gls{fpr95}$\downarrow$ & \gls{auroc}$\uparrow$  & \gls{fpr95}$\downarrow$ & \gls{auroc}$\uparrow$  &  \\
    \midrule
    EBO \cite{liuEnergybasedOutofdistributionDetection2020} & $70.0_{\pm 0.4}$ & $79.7_{\pm0.1}$ & $50.7_{\pm 0.9}$ & $85.1_{\pm 0.1}$ & $26.3_{\pm 2.3}$ & $92.5_{\pm 0.5}$ & $41.3_{\pm 1.9}$ & $90.8_{\pm 0.2}$ & $37.2_{\pm 1.7}$ & $89.1_{\pm 0.3}$ & $45.1_{\pm 14.7}$ & $87.4_{\pm 4.6}$\\
    ODIN \cite{hsuGeneralizedOdinDetecting2020a} & $74.9_{\pm 0.2}$ & $76.9_{\pm 0.1}$ & $60.1_{\pm 0.9}$ &  $83.3_{\pm 0.1}$ & $22.4_{\pm 1.8}$ &  $94.4_{\pm 0.4}$ & $43.0_{\pm 1.4}$ &  $90.7_{\pm 0.2}$ & $37.7_{\pm 1.0}$ &  $90.0_{\pm 0.2}$ & $47.6_{\pm 18.2}$ & $87.1_{\pm 6.2}$ \\
    MLS \cite{hendrycksScalingOutofdistributionDetection2022} & $70.0_{\pm 0.4}$ &  $80.1_{\pm 0.1}$ & $49.8_{\pm 0.8}$ &  $85.6_{\pm 0.1}$ & $25.0_{\pm 2.0}$ &  $93.1_{\pm 0.5}$ & $41.1_{\pm1.8}$ &  $90.6_{\pm 0.2}$ & $36.1_{\pm 1.3}$ &  $89.5_{\pm 0.3}$ & $44.4_{\pm 15.1}$ & $87.8_{\pm 4.5}$\\
    GradNorm \cite{huangImportanceGradientsDetecting2021a} &   $ 83.6_{\pm 0.9} $ & $ 71.3_{\pm 0.5}$ & $ 83.0_{\pm 0.3} $ & $ 73.3_{\pm 0.7}$ & $ 61.4_{\pm 3.0} $ & $ 86.1_{\pm 2.0}$ & $ 66.8_{\pm 3.5} $ & $ 86.1_{\pm 0.4}$ & $ 71.6_{\pm 0.4} $ & $ 80.4_{\pm 1.2}$ & $73.3_{\pm 8.8}$ & $79.4_{\pm 6.2}$\\
    ViM  \cite{wangViMOutofdistributionVirtuallogit2022} & $70.1_{\pm 0.5}$ & $74.0_{\pm 0.3}$ & $47.2_{\pm 1.1}$ & $83.3_{\pm 0.2}$ & $27.5_{\pm 0.4}$ & $91.0_{\pm 0.4}$ & $20.4_{\pm 0.2}$ & $94.6_{\pm 0.1}$ &  $34.1_{\pm 0.8}$ & $88.2_{\pm 0.2}$ & $39.9_{\pm 17.5}$ & $86.2_{\pm 7.1}$\\
    KNN \cite{sunOutofdistributionDetectionDeep2022} & 
   $72.3_{\pm 0.5}$ &  $77.0_{\pm 0.2}$ & $46.7_{\pm 0.8}$ &  $86.1_{\pm 0.1}$ & $24.4_{\pm 1.0}$ &  $94.0_{\pm 0.4}$ & $24.5_{\pm 0.2}$ &  $95.3_{\pm 0.02}$ & $33.4_{\pm 1.3}$ &  $90.1_{\pm 0.3}$ & $40.3_{\pm 18.0}$ & $88.5_{\pm 6.6}$\\
    CIDER \cite{mingHowExploitHyperspherical2023} & $75.5_{\pm  0.7}$ &				 $76.0_{\pm  2.4}$ &	    		 $44.7_{\pm  0.9}$ & $85.1_{\pm  1.1}$ & $26.5_{\pm  2.3}$ & $90.7_{\pm  2.1}$ & $31.5_{\pm  3.7}$ & $92.4_{\pm  1.4}$ & $32.5_{\pm  2.4}$ & $88.9_{\pm  1.6}$ & $42.1_{\pm  18.0}$ & $86.6_{\pm  5.8}$ \\  
    VOS \cite{duVOSLearningWhat2022} & $71.8_{\pm 0.9}$ & $78.6_{\pm 0.2}$ & $52.6_{\pm 0.9}$ & $84.4_{\pm 0.3}$ & $25.9_{\pm 0.8}$ & $92.4_{\pm 0.4}$ & $36.5_{\pm 0.5}$ & $91.7_{\pm 0.1}$ & $36.7_{\pm 1.1}$ & $89.0_{\pm 0.4}$ & $44.7_{\pm 16.0}$ & $87.2_{\pm 5.1}$ \\
    NPOS \cite{taoNonparametricOutlierSynthesis2022} & $72.9_{\pm 1.6}$ & $74.0_{\pm 0.5}$ & $49.1_{\pm 0.4}$ & $84.7_{\pm 0.2}$ & $22.8_{\pm 0.5}$ & $93.8_{\pm 0.2}$ & $18.1_{\pm 0.5}$ & $96.9_{\pm 0.1}$ & $28.2_{\pm 0.6}$ & $91.3_{\pm 0.2}$ & $38.2_{\pm 20.3}$ & $88.1_{\pm 8.1}$\\
    \midrule
    HOD ($\alpha=0$)  & $66.2_{\pm1.9}$ & $75.5_{\pm 0.5}$ & $44.5_{\pm 2.1}$ & $85.4_{\pm 0.2}$ & $19.8_{\pm 0.5}$ & $93.7_{\pm 0.3}$ & $28.3_{\pm 1.7}$ & $92.6_{\pm 0.3}$ & $\mathbf{0.0_{\pm 0.0}}$ & $\mathbf{100.0_{\pm 0.0}}$ & $31.8_{\pm 22.4}$ & $89.4_{\pm 8.4}$\\
    \bottomrule
    \end{tabular}%
    }
\end{table}

\paragraph{\textbf{Synthetic outliers do not benefit \gls{ood} detection in Hyperbolic space.}}
Our experiments provide evidence that incorporating synthetic outliers does not provide additional benefits for \gls{ood} detection using HOD. While such improvements were observed in Euclidean space with VOS \cite{duVOSLearningWhat2022} and NPOS \cite{taoNonparametricOutlierSynthesis2022}, our findings suggest that the inherent properties of Hyperbolic space are sufficient to achieve large class separation margins. It is also possible that the limitations of the Hyperbolic Wrapped Gaussian might hinder its ability to capture certain types of variation, particularly the potentially problematic nature of locally parallel principal axes in learning hierarchical representations \cite{cho2022a}.

\begin{figure}[t!]
    \centering
    \includegraphics{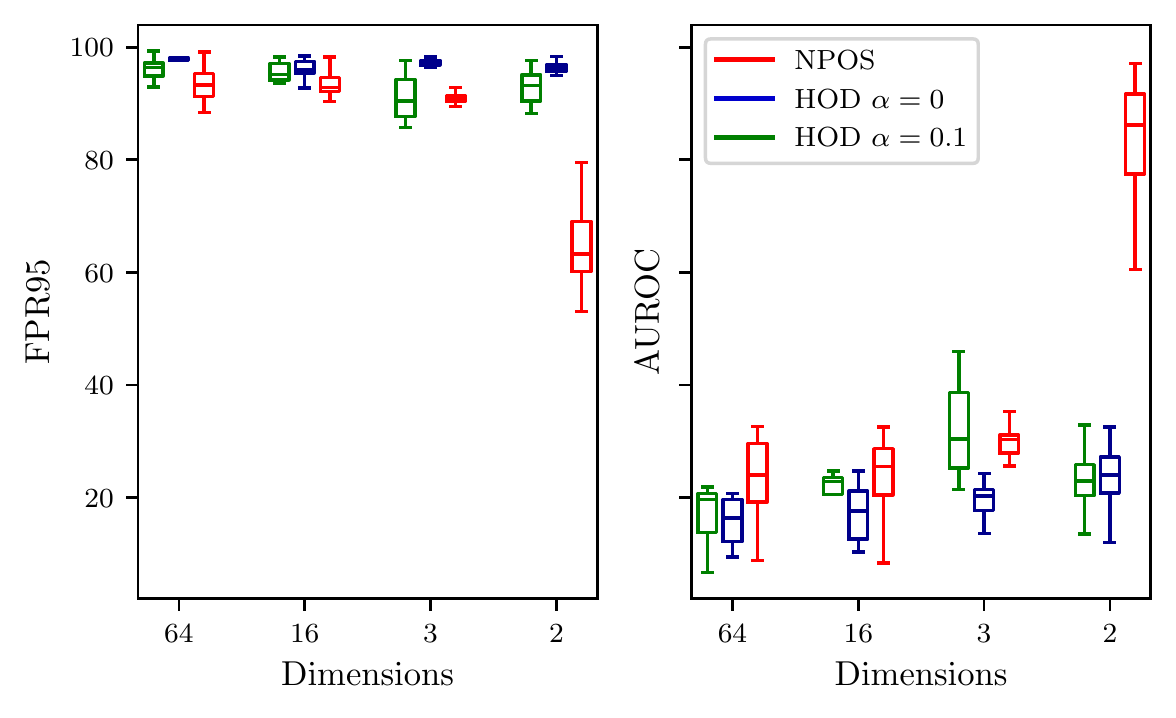}
    \caption{Boxplot illustrating \gls{ood} detection performance using {CIFAR-10} as the \gls{id} dataset across three distinct training runs. We present \gls{fpr95} and \gls{auroc} metrics upon reducing the dimensionality of the ResNet-18 projection layer. The Hyperbolic space maintains robustness to dimensionality reduction for \gls{ood} detection, which offers a promising solution for limited resource devices.}
    \label{fig:results_dimension_boxplot}
\end{figure}

\paragraph{\textbf{Low-dimensional out-of-distribution detection.}} Hyperbolic spaces leverage the representation space efficiently, enabling lower-dimensional embeddings for memory-constrained deployments, as shown in Kusupati \etal \cite{kusupati2022Matryoshka}. 

We evaluate NPOS and HOD, with and without synthetic outlier sampling, using projection layer dimensions of $\cbr{64, 16, 3, 2}$. \Cref{fig:results_dimension_boxplot} illustrates that Hyperbolic embeddings outperform NPOS in \gls{ood} detection at lower dimensions, making them a promising solution for resource-limited devices.

%% file: content/literature.tex
\section{Related Work}
\label{sec:literature}

\paragraph{\textbf{Out-of-Distribution detection.}} Research on \gls{ood} detection has witnessed significant advancements since the overconfidence phenomenon was discovered by Nguyen \etal \cite{nguyenDeepNeuralNetworks2015a}. These developments focused around two overarching methodologies: \textit{post-hoc} techniques and \textit{regularization-based} approaches.

Post-hoc methods perform \gls{ood} detection within pre-trained models through the development of
scoring functions. Score functions have been developed based on confidence \cite{bendaleOpenWorldRecognition2015, hendrycksBaselineDetectingMisclassified2018, liangEnhancingReliabilityOutofdistribution2018}, energy \cite{liuEnergybasedOutofdistributionDetection2020} distance \cite{leeSimpleUnifiedFramework2018, sehwagSSDUnifiedFramework2021, renSimpleFixMahalanobis2021, duSIRENShapingRepresentations2022, mingHowExploitHyperspherical2023}, gradients \cite{huangImportanceGradientsDetecting2021a, gonzalez-jimenezSANOScoreBasedDiffusion2023}, and Bayesian statistics \cite{galDropoutBayesianApproximation2016a, lakshminarayananSimpleScalablePredictive2017a, 
gonzalez-jimenezRobustTLossMedical2023,
maddoxSimpleBaselineBayesian2019, wenBatchEnsembleAlternativeApproach2020, kristiadiBeingBayesianEven2020}.
However, those methodologies are based on Euclidean models which struggle to capture intricate hierarchical structures often found in natural images. Some works suggested using the Hyperbolic space for \gls{ood} detection \cite{vanspenglerPoincarResNet2023, guoClippedHyperbolicClassifiers2022}, yet they have fallen short of achieving competitive results compared to the state-of-the-art. HOD address this gap by proposing a combination of supervised contrastive loss and Hyperbolic \gls{knn} which proves effective.

On a different line, regularization-based approaches, which focus on training-time adjustments to the model to foster better discrimination between \gls{id} and \gls{ood} data \cite{bevandicDiscriminativeOutofdistributionDetection2018, malininPredictiveUncertaintyEstimation2018, geifmanSelectiveNetDeepNeural2019, heinWhyReluNetworks2019a, meinkeNeuralNetworksThat2020a, jeongOODMAMLMetalearningFewshot2020, yangSemanticallyCoherentOutofdistribution2021a}. 
These methods often involve regularization techniques aimed at encouraging the model to produce lower confidence \cite{leeTrainingConfidencecalibratedClassifiers2018, hendrycksDeepAnomalyDetection2019} or higher energy score \cite{duVOSLearningWhat2022, liuEnergybasedOutofdistributionDetection2020, katz-samuelsTrainingOODDetectors2022, mingPOEMOutofdistributionDetection2022}. 

However, these methods require the availability of auxiliary \gls{ood} datasets. Notably, VOS \cite{duVOSLearningWhat2022} and NPOS \cite{taoNonparametricOutlierSynthesis2022} have pioneered the synthesis of outliers, providing a flexible alternative to using real \gls{ood} data for model regularization. Despite these advancements, challenges persist, particularly in estimating the \gls{id} to sample from without distributional assumptions. We partially address this issue with a new framework to sample the synthetic outliers in the Hyperbolic space. Due to the learned hierarchical structure, it is indeed possible to sample from those \gls{id} embeddings that are more uncertain or encompass broader concepts. This offers a principled strategy to obtain synthetic outliers.

\paragraph{\textbf{Hyperbolic representation.}} Hyperbolic space has gained significant attention in \gls{dl}, particularly for tasks involving hierarchical structures, \ie, tree-like structures and taxonomies \cite{ganeaHyperbolicNeuralNetworks2018, lawLorentzianDistanceLearning2019, nickelPoincarEmbeddingsLearning2017}, text representations \cite{desaiHyperbolicImagetextRepresentations2024, cimpoiDescribingTexturesWild2014, zhuHyperTextEndowingFastText2020}, and graphs \cite{bachmannConstantCurvatureGraph2020, chamiHyperbolicGraphConvolutional2019, daiHyperbolictoHyperbolicGraphConvolutional2021}.
Previous research has underscored the potential of adopting a hierarchical perspective for \gls{ood} detection. For instance, Flaborea \etal \cite{flaborea2022certain} use the Hyperbolic space to provide an uncertainty score for time series. Li \etal
\cite{li_hyperbolic_anomaly_detection_CVPR} obtain an anomaly map in industrial images by calculating the difference between image embeddings in the hyperbolic space.
Similarly, the distance to the origin in the Poincar\'e ball has been highlighted to be a good uncertainty metric used for image segmentation \cite{ghadimiatighHyperbolicImageSegmentation2022}, classification \cite{khrulkovHyperbolicImageEmbeddings2020} and few-shot learning \cite{anvekar2023gpr}.

The use of softmax and energy score have been explored in the Hyperbolic space for \gls{ood} detection too \cite{vanspenglerPoincarResNet2023, guoClippedHyperbolicClassifiers2022}.
However, results were found not to be competitive with Euclidean methodologies. Our work addresses the limitations encountered by previous approaches by leveraging the Hyperbolic structure to learn a clear boundary between in-distribution and out-of-distribution data, and achieving state-of-the-art performance for \gls{ood} detection.

%% file: content/limitations.tex
\section{Limitations}
\label{sec:limitations}
In this work, we focused on demonstrating that Hyperbolic \gls{dl} achieves competitive results for \gls{ood} detection with an appropriate combination of training objective and \gls{ood} score, even for small embedding dimensions.
To this end, we considered typical benchmarks based on natural image datasets. CIFAR-10 was chosen for cheap hyperparameter tuning and ablation studies despite its flat semantic structure. CIFAR-100 and ImageNet-200 were instead selected because they exhibit hierarchical concepts with manageable complexity. Further investigation on large-scale datasets, deep taxonomies, and specific domains is needed, as generalization to complex cases such as ICD-10 diagnosis on large medical image collections is far from trivial.
Similarly, even if we include a study on model depth in the supplementary material, other architectures may feature different behavior. In addition, recent fully Hyperbolic architectures \cite{FullyHyperbolicConvolutional2023, mettesHyperbolicDeepLearning2023} may bring additional benefits. However, training such architectures can be complex and prone to numerical instability \cite{mishneNumericalStabilityHyperbolic2023}. Our approach maps embeddings from the last layer to the Lorentz space avoding these instabilities and is superior from the perspective of optimization, in contrast to the Poincaré ball model \cite{mishneNumericalStabilityHyperbolic2023}. Although the Poincaré ball model may offer broader representation capacity and potentially improved results, its exploration remains outside the scope of this paper and is reserved for future research.
In terms of computational complexity, our approach effectively modifies the Euclidean training objective by lifting the representation of the final layer to the hyperbolic space. This has to be gauged against VOS \cite{duVOSLearningWhat2022} and NPOS \cite{taoNonparametricOutlierSynthesis2022}, which invest computation to dynamically sample virtual outliers, or against CIDER \cite{mingHowExploitHyperspherical2023} and HYPO \cite{bai2024hypo}, which consider modified training objectives similar to us. When we do not consider the generation of virtual outliers in Hyperbolic space, our runtime is very close to its Euclidean counterpart as typical for this class of methods.

%% file: content/conclusions.tex
\section{Conclusions}
\label{sec:conclusions}

Traditional \gls{ood} detection methods have limitations, often suffering from overconfidence or reliance on external datasets. The introduction of Hyperbolic geometry as an alternative to the Euclidean space offers promising avenues for addressing these challenges. We developed an effective framework for \gls{ood} detection, HOD, by leveraging the hierarchical representation capabilities of Hyperbolic space.

Our empirical evaluations demonstrate substantial performance improvements, as HOD significantly outperforms competing state-of-the-art methods in terms of \gls{fpr95} and \gls{auroc} on {CIFAR-10} and {CIFAR-100} benchmarks, and matches competing scores on ImageNet-200. We also introduced a novel, principled strategy for synthesizing outliers in Hyperbolic space to avoid the need for auxiliary datasets, and found evidence that it may be redundant within HOD. Additionally, we showed that the Hyperbolic space holds potential for real-world deployment, especially in resource-constrained environments.

The implications of our findings extend beyond \gls{ood} detection, offering opportunities for practical applications across various industrial domains. By bridging the gap between Euclidean and Hyperbolic space, our research paves the way for more robust and efficient \gls{dl} models capable of handling unforeseen data distributions.

%% file: content/appendix.tex
\clearpage
\begin{center}
	\Large{\textbf{Hyperbolic Metric Learning for Visual Outlier Detection}}
\end{center}

\section{Ablation Synthetic Outliers}

\begin{table}
	\caption{Effect of the variance $\sigma$ for CIFAR-10 \gls{id} and \gls{ood} validation data.}
	\label{tab:train_sigma_ablation}%
	\resizebox{\textwidth}{!}{
		\begin{tabular}{cccccccccc}
			\toprule
			\multirow{3}[4]{*}{$\sigma$} & \multicolumn{8}{c}{$D^{\text{val}}_{\text{out}}$}                                                                                                                                                                                                  \\
			\cmidrule{2-10}
			                             & \multicolumn{2}{c}{LSUN}                          & \multicolumn{2}{c}{LSUN Resize} & \multicolumn{2}{c}{iSUN} & \multicolumn{2}{c}{Average}                                                                                                       \\
			\cmidrule{2-10}
			                             & \gls{fpr95}$\downarrow$                           & \gls{auroc}$\uparrow$           & \gls{fpr95}$\downarrow$  & \gls{auroc}$\uparrow$       & \gls{fpr95}$\downarrow$ & \gls{auroc}$\uparrow$ & \gls{fpr95}$\downarrow$ & \gls{auroc}$\uparrow$ & \\
			\midrule
			0.01                         & 10.15                                             & 98.48                           & 3.50                     & 99.23                       & 1.96                    & 99.50                 & 5.20                    & 99.07                   \\
			0.03                         & 11.87                                             & 97.14                           & 3.10                     & 99.25                       & 2.58                    & 99.28                 & 5.85                    & 98.56                   \\
			0.08                         & 10.59                                             & 96.69                           & 9.88                     & 98.74                       & 8.32                    & 99.11                 & 9.60                    & 98.18                   \\
			0.1                          & 12.74                                             & 97.08                           & 10.76                    & 99.35                       & 10.44                   & 99.46                 & 11.31                   & 98.63                   \\
			\bottomrule
		\end{tabular}%
	}
\end{table}

\paragraph{\textbf{Effect of the variance $\sigma$}.} The parameter $\sigma$ regulates the spread of synthetic outliers around the seed uncertain \gls{id} embeddings.
By varying $\sigma$ across different values $\cbr{0.01, 0.02, 0.03, 0.06, 0.08, 0.1}$, the effect on \gls{ood} detection performance was systematically analyzed.
It was observed in \cref{tab:train_sigma_ablation} that the performance of the method was relatively stable under moderate variances.

\begin{table}
	\caption{Effect of regularization weight $\alpha$ for CIFAR-10 \gls{id} and \gls{ood} validation data.}
	\resizebox{\textwidth}{!}{
		\begin{tabular}{cccccccccc}
			\toprule
			\multirow{3}[4]{*}{$\alpha$} & \multicolumn{8}{c}{$D^{\text{val}}_{\text{out}}$}                                                                                                                                                                                                  \\
			\cmidrule{2-10}
			                             & \multicolumn{2}{c}{LSUN}                          & \multicolumn{2}{c}{LSUN Resize} & \multicolumn{2}{c}{iSUN} & \multicolumn{2}{c}{Average}                                                                                                       \\
			\cmidrule{2-10}
			                             & \gls{fpr95}$\downarrow$                           & \gls{auroc}$\uparrow$           & \gls{fpr95}$\downarrow$  & \gls{auroc}$\uparrow$       & \gls{fpr95}$\downarrow$ & \gls{auroc}$\uparrow$ & \gls{fpr95}$\downarrow$ & \gls{auroc}$\uparrow$ & \\
			\midrule
			0.1                          & 10.65                                             & 98.22                           & 7.67                     & 99.12                       & 0.95                    & 99.59                 & 6.42                    & 98.98                   \\
			0.15                         & 10.78                                             & 93.08                           & 3.15                     & 98.97                       & 1.41                    & 99.22                 & 5.11                    & 97.09                   \\
			0.5                          & 10.73                                             & 95.93                           & 6.65                     & 98.65                       & 3.24                    & 98.98                 & 6.87                    & 97.85                   \\
			1                            & 14.13                                             & 95.44                           & 5.69                     & 98.92                       & 5.93                    & 98.84                 & 8.58                    & 97.73                   \\
			\bottomrule
		\end{tabular}%
	}
	\label{tab:train_alpha_ablation}%
\end{table}

\paragraph{\textbf{Effect of regularization weight $\alpha$.}} The weight parameter $\alpha$ in the training objective plays a crucial role in determining the influence of virtual outliers. We investigate the impact of varying $\alpha$ values on \gls{ood} detection by training models with different weights, specifically $\alpha \in \cbr{0.0, 0.1, 0.2, 0.5, 1}$. Using CIFAR-10 as $D_\text{in}$ data and evaluating on $D_{\text{out}}^{\text{val}}$, we observe notable differences in performance. For instance, employing a moderate weighting such as $\alpha=0.1$ leads to a significant enhancement in \gls{ood} detection performance, as seen in \cref{tab:train_alpha_ablation}.

\begin{table}
	\caption{Effect of starting sampling synthetic outliers at different iterations for CIFAR-10 \gls{id} and \gls{ood} validation data.}
	\resizebox{\textwidth}{!}{
		\begin{tabular}{cccccccccc}
			\toprule
			\multirow{3}[4]{*}{Iteration} & \multicolumn{8}{c}{$D^{\text{val}}_{\text{out}}$}                                                                                                                                                                                                  \\
			\cmidrule{2-10}
			                              & \multicolumn{2}{c}{LSUN}                          & \multicolumn{2}{c}{LSUN Resize} & \multicolumn{2}{c}{iSUN} & \multicolumn{2}{c}{Average}                                                                                                       \\
			\cmidrule{2-10}
			                              & \gls{fpr95}$\downarrow$                           & \gls{auroc}$\uparrow$           & \gls{fpr95}$\downarrow$  & \gls{auroc}$\uparrow$       & \gls{fpr95}$\downarrow$ & \gls{auroc}$\uparrow$ & \gls{fpr95}$\downarrow$ & \gls{auroc}$\uparrow$ & \\
			\midrule
			1000                          & 11.63                                             & 96.34                           & 5.30                     & 99.04                       & 3.28                    & 99.22                 & 6.73                    & 97.2                    \\
			5000                          & 11.19                                             & 93.15                           & 3.45                     & 98.65                       & 4.89                    & 98.32                 & 6.5                     & 96.7                    \\
			10,000                        & 10.75                                             & 98.88                           & 15.32                    & 98.10                       & 13.84                   & 98.31                 & 13.3                    & 98.4                    \\
			15,000                        & 11.30                                             & 97.95                           & 15.45                    & 98.05                       & 13.93                   & 98.96                 & 13.6                    & 98.3                    \\
			\bottomrule
		\end{tabular}%
	}
	\label{tab:iteration_ablation_study}%
\end{table}

\paragraph{\textbf{Choosing the initial iteration for sampling}.} In \cref{tab:iteration_ablation_study}, we examine the impact of selecting the starting iteration for introducing synthetic outliers during training. Incorporating synthetic outliers in the latter phases of training results in a slightly inferior out-of-distribution detection performance. This is attributed to insufficient iterations for the network to adequately regularize. Conversely, introducing regularization during the mid-training phase yields more favorable performance.

\clearpage

\section{Ablation on Encoder Depth}
\label{app:ablationencoderdepth}

\begin{figure}
	\centering
	\includegraphics[width=.85\textwidth]{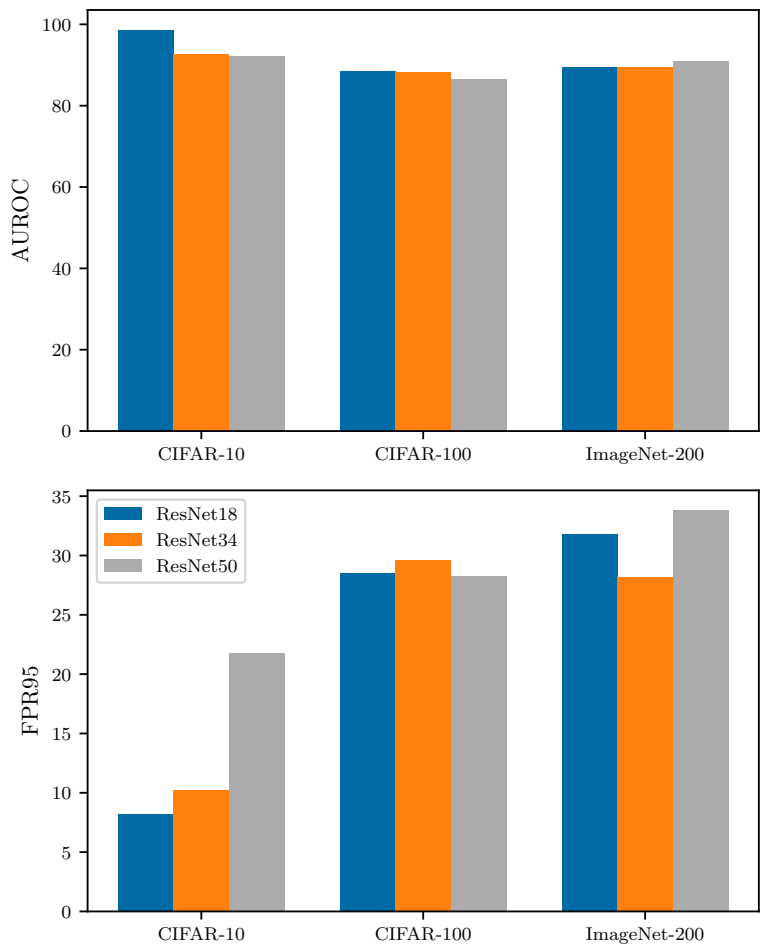}
	\caption{Average OOD detection performance using CIFAR-10/100 and ImageNet-200 as $D_\text{id}$ and with ResNet-18, ResNet-34 and ResNet-50 backbones.}
	\label{fig:ablation-backbones}
\end{figure}

Our method applies to deeper encoders, as it obtains features in the hyperbolic space with changes only in the last layer. \Cref{fig:ablation-backbones} shows \gls{auroc} and \gls{fpr95} scores for ResNet-34 and ResNet-50 in the three considered datasets, demonstrating that our approach is robust to different encoders.

\clearpage
\section{Visualization of the Outliers in the Hyperbolic space}
\begin{figure}[hbt!]
	\begin{subfigure}{.4\linewidth}
		\includegraphics[width=\linewidth]{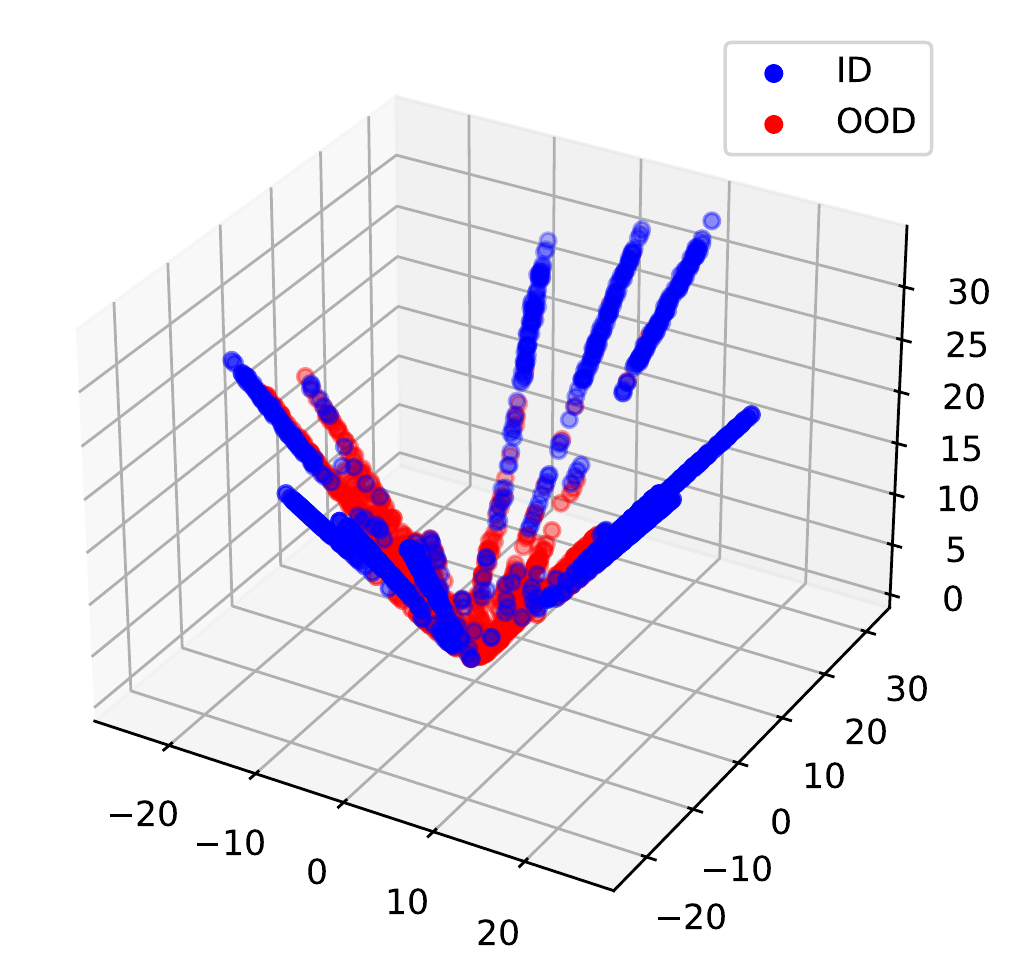}
		\caption{SVHN}
	\end{subfigure}\hfill 
	\begin{subfigure}{.4\linewidth}
		\includegraphics[width=\linewidth]{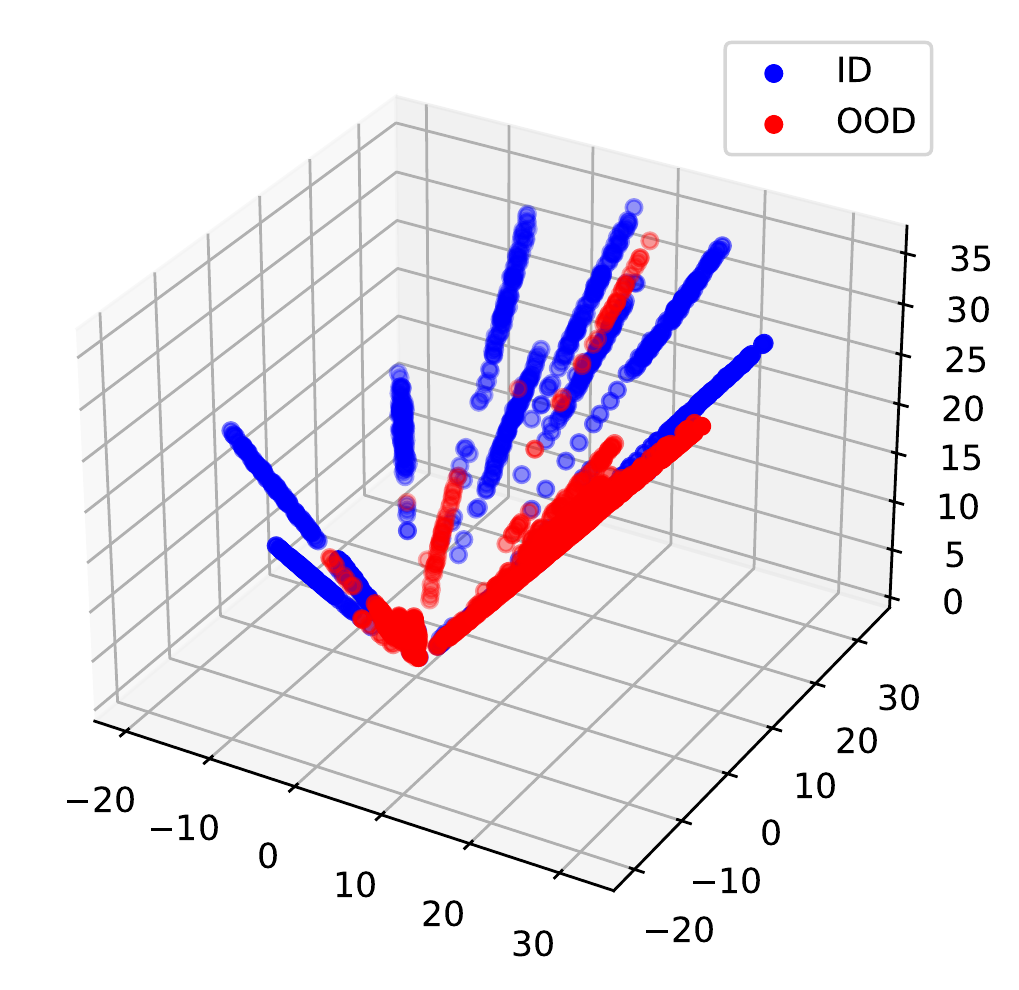}
		\caption{MNIST}
	\end{subfigure}

	\medskip 
	\begin{subfigure}{.4\linewidth}
		\includegraphics[width=\linewidth]{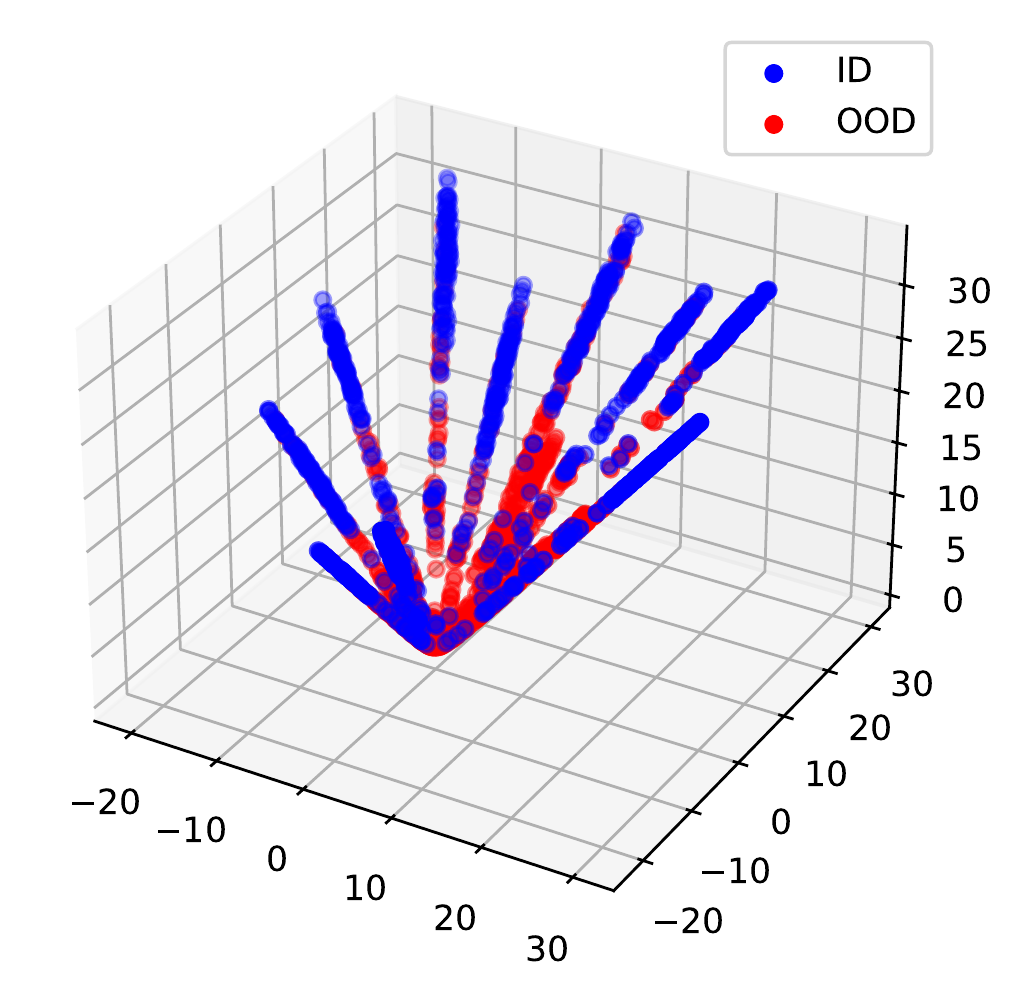}
		\caption{Places365}
		\label{velcomp}
	\end{subfigure}\hfill 
	\begin{subfigure}{.4\linewidth}
		\includegraphics[width=\linewidth]{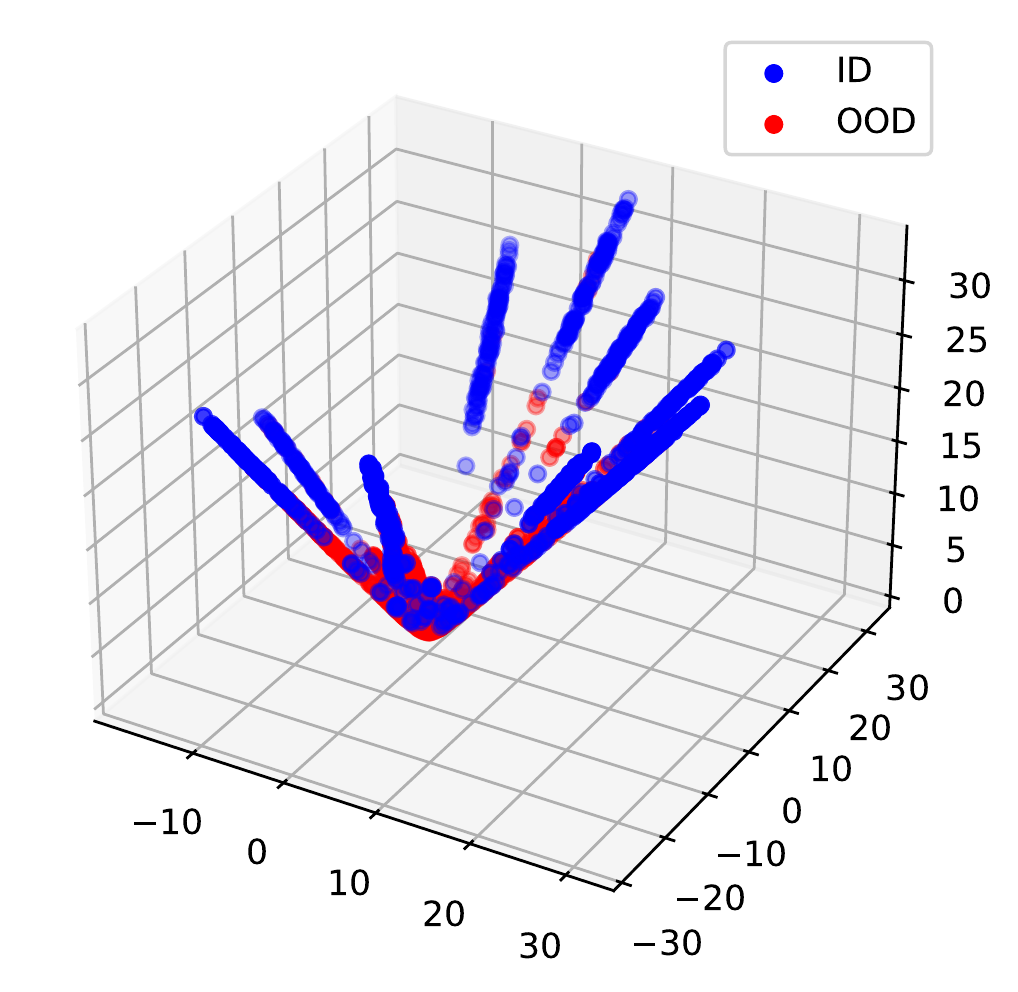}
		\caption{Textures}
		\label{estcomp}
	\end{subfigure}

	\caption{UMAP visualization of the \gls{id} and \gls{ood} embeddings in the Hyperbolic space. The blue points denote the embeddings of CIFAR-10 and the red points denotes the embeddings of the \gls{ood} datasets.}
	\label{fig:embeddings_ood_umap}
\end{figure}

In this section, we employ UMAP to visualize the embeddings of the \gls{ood} datasets within the Hyperbolic space. \Cref{fig:embeddings_ood_umap} shows that the embeddings corresponding to the in-distribution data are clustered and distinctly separated. While it may appear that there is some overlap between \gls{id} and \gls{ood} embeddings, it's important to note that the distances between them are significant, owing to the exponential properties in the Hyperbolic space.